\pdfoutput=1
\PassOptionsToPackage{prologue,dvipsnames}{xcolor}
\documentclass[11pt]{article}

\usepackage[final]{acl}

\usepackage{times}
\usepackage{latexsym}

\usepackage[T1]{fontenc}

\usepackage[utf8]{inputenc}

\usepackage{microtype}

\usepackage{inconsolata}

\usepackage{graphicx}

\usepackage{times}
\usepackage{epsfig}
\usepackage{amsmath, amssymb, amsfonts}
\usepackage{enumitem}
\usepackage{float}
\usepackage{textcomp}
\usepackage{graphicx}
\usepackage{subcaption}
\usepackage{booktabs} %
\usepackage{multirow} %
\usepackage{array} %
\usepackage{tabularx}
\usepackage{longtable}
\usepackage{xspace}
\usepackage{adjustbox}

\usepackage{lipsum}
\usepackage{bbm}
\usepackage{stmaryrd}
\usepackage{makecell}
\usepackage{courier}
\usepackage{bbm}
\usepackage{algorithm, algpseudocode}
\usepackage{setspace}
\usepackage{threeparttable}
\usepackage{cancel}
\usepackage{latexsym}
\usepackage{dirtytalk}
\usepackage{csquotes}
\usepackage{pgfplots}
\usepackage{pifont}
\usepackage[dvipsnames]{xcolor}
\usepackage{geometry}
\usepackage{adjustbox}
\usepackage{lineno}

\usepackage{sections/package}
\newcommand{\cmark}{\ding{51}}
\newcommand{\xmark}{\ding{55}}
\definecolor{mygreen}{RGB}{34,139,34}
\definecolor{myred}{RGB}{178,34,34}

\newcommand{\modelname}{EgoSpeak\xspace}
\newcommand{\frameworkname}{EgoSpeak\xspace}

\title{\modelname: Learning When to Speak\\ for Egocentric Conversational Agents in the Wild}

\author{
Junhyeok Kim$^{\clubsuit}$ \quad
Min Soo Kim$^{\clubsuit}$ \quad
Jiwan Chung$^{\clubsuit}$ \quad
Jungbin Cho$^{\clubsuit}$\\
\textbf{Jisoo Kim}$^{\clubsuit}$ \quad
\textbf{Sungwoong Kim}$^{\clubsuit}$ \quad
\textbf{Gyeongbo Sim}$^{\diamondsuit}$ \quad
\textbf{Youngjae Yu}$^{\clubsuit}$\\
\small{$\clubsuit$ Yonsei University} \quad
\small{$\diamondsuit$ Multimodal AI Lab., NC Research, NCSOFT Corporation} \\
\texttt{junhyeok@yonsei.ac.kr}
}

\begin{document}
\maketitle
\begin{abstract}
Predicting when to initiate speech in real-world environments remains a fundamental challenge for conversational agents. We introduce \frameworkname, a novel framework for real-time speech initiation prediction in egocentric streaming video. By modeling the conversation from the speaker’s first-person viewpoint, \frameworkname is tailored for human-like interactions in which a conversational agent must continuously observe its environment and dynamically decide when to talk.

Our approach bridges the gap between simplified experimental setups and complex natural conversations by integrating four key capabilities: (1) first-person perspective, (2) RGB processing, (3) online processing, and (4) untrimmed video processing. We also present YT-Conversation, a diverse collection of in-the-wild conversational videos from YouTube, as a resource for large-scale pretraining. Experiments on EasyCom and Ego4D demonstrate that \frameworkname outperforms random and silence-based baselines in real time. Our results also highlight the importance of multimodal input and context length in effectively deciding when to speak. Code and data are available at \href{https://jun297.github.io/EgoSpeak/}{website}. 
\end{abstract}

\section{Introduction}\label{sec:introduction}

Human-like conversational agents have long been a key objective in artificial intelligence. A critical aspect of human conversation is not only understanding what to say but also when to say it—often framed as the study of turn-taking \cite{duncan1972some}. While most are designed under simplified assumptions where turn boundaries are well-defined or where only audio-based cues are available, real-world conversations can be highly fluid, with overlapping speech, unclear speaker roles, and frequent interruptions \cite{skantze2017towards, skantze2021turnreview}.

To address these complexities, we introduce \frameworkname, a framework that predicts when an agent should begin speaking based on egocentric streaming video. Concretely, \frameworkname models speech initiation from the first-person perspective of the camera wearer, capturing exactly what the agent sees at each moment in real time. 
Unlike a third-person or fixed camera view, the egocentric perspective is especially relevant 
for real-world conversational agents such as social robots that must decide on the fly whether to speak or remain silent. By leveraging the camera wearer’s immediate visual context (e.g., facing another person, noticing body language or gaze direction), \frameworkname can more naturally detect subtle cues that signal an appropriate moment to start speaking. This is particularly crucial for a real-world agent that must not only process inputs in real time, but also respond autonomously in dynamic, multi-speaker environments 
to appear natural and engaging.

\begin{figure}[t]
\centering
\includegraphics[trim={0 0cm 0 0},width=1.0\linewidth]{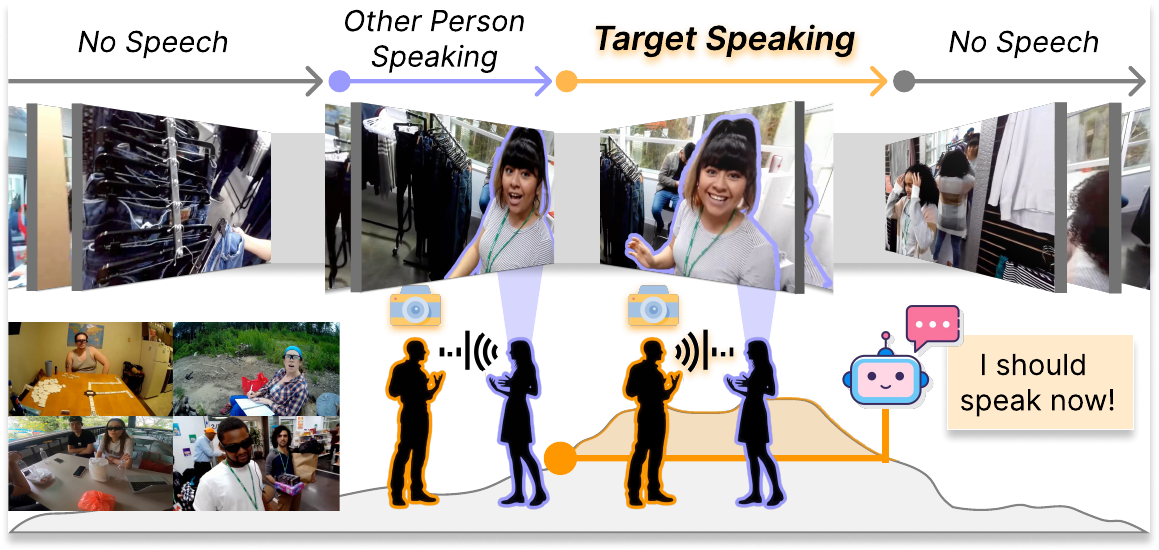}

\caption{\frameworkname\ models speech initiation in real time from the camera wearer’s (camera icon) egocentric video stream, mirroring how a real-world agent would perceive and engage in dynamic, multi-speaker environments.}

\label{fig:new_teaser}
\vspace*{-1.5em}
\end{figure}

\begin{figure*}[t]
\centering
\includegraphics[width=0.99\linewidth]{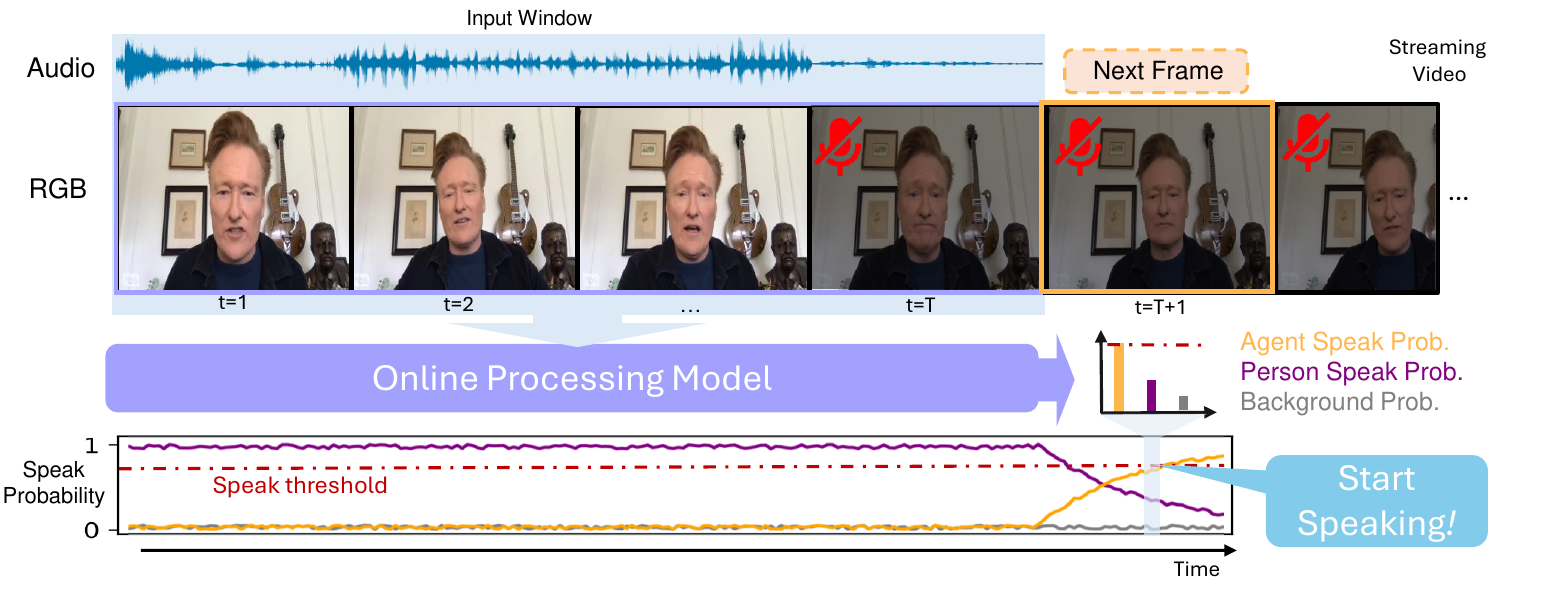}
\caption{Overview of the \frameworkname framework. At each time step, the model processes an untrimmed egocentric video and audio stream, classifying them in real time into three categories: background (no speech), other person speaking, and target speaker (camera wearer) speaking. These probabilities are visualized at the bottom, where the model anticipates near-future frames and enables proactive speech initiation for conversational agents.}

\vspace*{-1em} %
\label{fig:method_desc}
\end{figure*}

\frameworkname incorporates four key capabilities: (1) first-person perspective: aligns closely with real-world interactions for conversational agents, (2) RGB feature processing: handles scenarios where audio or non-verbal cues may be unreliable, (3) dynamic real-time turn-taking: enables more natural and fluid conversations, and (4) continuous untrimmed video stream processing: captures periods of silence and sporadic interactions. These four collectively enable \frameworkname to handle the complexities of real-world conversations more effectively than previous methods (see \Cref{tab:task_position}). \Cref{fig:method_desc} provides an overview of our real-time pipeline, illustrating how \frameworkname processes continuous video streams to decide when to speak. \frameworkname outputs a continuous speak-probability that a conversational agent can leverage in real time (e.g., by triggering speech once the probability surpasses a threshold).

We validate \frameworkname on two distinct datasets: EasyCom and Ego4D, demonstrating its effectiveness across various conversational contexts. Additionally, we introduce the YT-Conversation dataset, a collection of in-the-wild conversation videos including interviews and casual conversations from YouTube, designed for scalable pretraining.  

By addressing the critical challenge of when to speak in a natural, human-like manner, \frameworkname advances the field of conversational AI, offering a robust solution for dynamic, intermittent conversations with varying numbers of speakers.  

In summary, our key contributions are:
\begin{enumerate}[nosep, leftmargin=*]

\item \frameworkname, a novel framework for speech initiation prediction from egocentric streaming video in real time.
\item YT-Conversation, a large-scale corpus of in-the-wild conversational videos, suitable for 
pretraining multimodal turn-taking models.
\item Experimental results on EasyCom and Ego4D demonstrate effectiveness in real-world scenarios and provide a comprehensive analysis of the role of multimodal inputs and context length. 

\end{enumerate}

\section{Related Works}\label{sec:relatedworks}
\paragraph{Turn-taking.}
Turn-taking research has evolved from simple audio-based models \cite{duncan1972some, khouzaimi2015optimising} to sophisticated multimodal approaches \cite{maier2017LSTMtowards, lee2023multimodal, mizuno2023next, kurata2023multimodal}. Early offline methods, which process entire clips, often result in unnatural pauses. This prompted the development of continuous (online) methods \cite{skantze2017towards, ekstedt2022voice, li2022can}, including recent multimodal models incorporating non-verbal cues \cite{onishi2023multimodalvap}. However, these approaches typically rely on controlled dyadic conversations, limiting real-world applicability. 
\frameworkname addresses these limitations by adopting a first-person perspective, processing both RGB and audio features, and handling untrimmed video streams, aiming to better align turn-taking models with the complexities of natural conversations.

\begin{table}[hb!]  %
\centering
\begin{minipage}{\columnwidth} 
\centering
\begin{adjustbox}{width=\columnwidth}
\begin{tabular}{lcccc}
\toprule
 & Egocentric & RGB & Online & Untrimmed \\
\midrule
~\citet{skantze2017towards} & \textcolor{myred}{\xmark} & \textcolor{myred}{\xmark} & \textcolor{mygreen}{\cmark} & \textcolor{myred}{\xmark} \\
~\citet{ekstedt2022voice} & \textcolor{myred}{\xmark} & \textcolor{myred}{\xmark} & \textcolor{mygreen}{\cmark} & \textcolor{myred}{\xmark} \\
~\citet{li2022can} & \textcolor{myred}{\xmark} & \textcolor{myred}{\xmark} & \textcolor{mygreen}{\cmark} & \textcolor{myred}{\xmark} \\
~\citet{yang2022gated} & \textcolor{myred}{\xmark} & \textcolor{myred}{\xmark} & \textcolor{myred}{\xmark} & \textcolor{myred}{\xmark} \\
~\citet{kurata2023multimodal} & \textcolor{myred}{\xmark} & \textcolor{mygreen}{\cmark} & \textcolor{myred}{\xmark} & \textcolor{myred}{\xmark} \\
~\citet{lee2023multimodal} & \textcolor{myred}{\xmark} & \textcolor{mygreen}{\cmark} & \textcolor{myred}{\xmark} & \textcolor{myred}{\xmark} \\
~\citet{mizuno2023next} & \textcolor{myred}{\xmark} & \textcolor{myred}{\xmark} & \textcolor{myred}{\xmark} & \textcolor{myred}{\xmark} \\
~\citet{onishi2023multimodalvap} & \textcolor{myred}{\xmark} & \textcolor{myred}{\xmark} & \textcolor{mygreen}{\cmark} & \textcolor{myred}{\xmark} \\
~\citet{fatan20243m} & \textcolor{mygreen}{\cmark} & \textcolor{mygreen}{\cmark} & \textcolor{myred}{\xmark} & \textcolor{myred}{\xmark} \\
\midrule
\textbf{EgoSpeak (Ours)} & \textcolor{mygreen}{\cmark} & \textcolor{mygreen}{\cmark} & \textcolor{mygreen}{\cmark} & \textcolor{mygreen}{\cmark} \\
\bottomrule
\end{tabular}
\end{adjustbox}
\caption{Comparison of EgoSpeak with existing utterance initiation and turn-taking methods. EgoSpeak uniquely addresses all four key aspects of real-world conversational dynamics.}
\label{tab:task_position}
\end{minipage}
\end{table}

\paragraph{Online Processing.}
Online systems, which predict based only on past and present information continuously, have gained popularity in real-time applications across various fields from computer vision to speech \cite{fan2018online, de2016online, kang2021cag, bewley2016simple, rettig2019online, miao2020transformer}.  \frameworkname applies this approach to turn-taking, enabling real-time prediction of speech initiation points in natural conversations without relying on future information. This capability allows \frameworkname to adapt to dynamic conversational scenarios, making it more suitable for real-world interactions.

\section{EgoSpeak Framework}
\label{sec:method}

\subsection{Framework Overview}
\frameworkname is designed for in-the-wild conversational agents, building on the challenges discussed in \Cref{sec:introduction}, where "in-the-wild" refers to real-world conditions outside controlled environments with unpredictable variables and numerous influencing factors.

\frameworkname is grounded in the intuition that, in the egocentric video, the camera wearer’s speaking moments naturally serve as cues for speech initiation. By predicting these moments from the agent’s perspective, our framework learns natural turn-taking behavior, identifying when to speak even after long silences. Moreover, by anticipating these moments in advance, \frameworkname effectively mirrors human turn-taking, deciding when to begin speaking as a real-world agent would. To achieve this, we train the model with a cross-entropy objective, akin to next-token prediction in language modeling, since it must anticipate speaking before the camera wearer actually speaks.

\subsection{Task Definition}
Guided by this intuition, we formulate the problem of predicting the target speaker's speech in an egocentric streaming video, where the camera wearer is naturally identified as the target speaker. Given the real-time nature of the stream, \frameworkname only analyzes information available up to the current moment. This design allows our system to capture the continuously unfolding context and prepare a speech onset before a turn-shift occurs in complex, dynamic conversations.

Formally, let $X_t = [x_1, \ldots, x_t]$ be an online stream up to timestep $t$, where 
each $x_i$ can include multiple modalities $x^m_i$ including visual frames $x^v_i$ or auditory signals $x^a_i$. We transform each $x^m_i$ into a representation $z^m_i$ via off-the-shelf feature extractors, concatenating them into $z_i$. Next, we define a temporal window $Z_{t-L+1,t} = [z_{t-L+1}, \ldots, z_t]$ of length $L$. 
Given an anticipation length $\alpha$, the model performs a three-way classification (background / target speaker speaking / other speaking) for the future range $t+1$ to $t+\alpha$. This anticipatory modeling gives the system extra time to prepare responses, rather than reacting only after a silence threshold. The final model output is a probability tensor of shape $[\alpha, 3]$, where the dimensions correspond to the anticipated future timesteps and the three classes, respectively.

\paragraph{Prediction vs Detection.}
A naive approach for determining when to speak is detection which occurs based on silence threshold. However, detection offers an inadequate response time of only 200ms for listeners. A psycholinguistic study \cite{levinson2015timing} estimates that actual response time ranges from 600 to 1500ms, as humans begin preparing their responses while the other person is still speaking. Additionally, turn-shifts often occur as overlapping without any gaps \cite{skantze2021turnreview}. The prediction will give conversational systems more time to generate reactions and enable human-like conversation.

\begin{figure}[t]
\centering
\includegraphics[trim={0 0cm 0 0},width=1\linewidth]{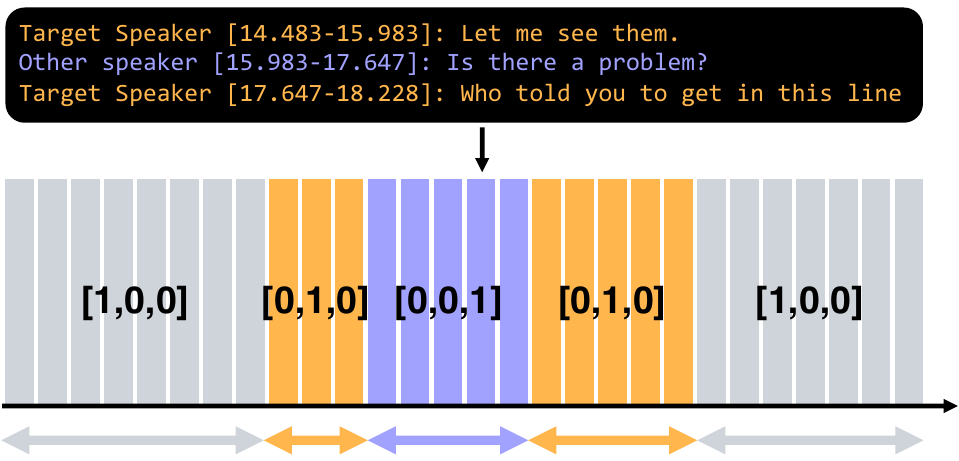}
\caption{
Converting Transcript to Per-Frame Labels. Colors indicate: gray - background, orange - target speaker speaking, purple - other speaker speaking. Labels are one-hot encoded for classification.
}
\vspace*{-1.5em}
\label{fig:data_annotation}
\end{figure}

\paragraph{Frame-level Speech Labeling.}

\Cref{fig:data_annotation} illustrates how transcript timestamps convert into 
per-frame, one-hot encoded labels. As our framework requires per-frame speech labels which are expensive to annotate, we developed a method to convert transcript annotations from egocentric videos into per-frame speech classification labels. At each timestep $t$, we label the datapoint $x_t$ as \textit{target speaker speaking} if the camera wearer is speaking, \textit{other person speaking} if others are speaking, and \textit{no speech} otherwise.

\subsection{YT-Conversation: Dataset for Multimodal Conversation Pretraining}
Existing turn-taking resources often stem from controlled laboratory setups or video calls, which are expensive to annotate and capture only a fraction of the complexity found in real-world interactions, limiting scalability. To address this gap, we introduce YT-Conversation, a novel dataset derived from diverse YouTube content including interviews, podcasts, and casual dialogues.

While YT-Conversation is not fully egocentric, it offers realistic face-to-face and multi-person interactions that can effectively transfer to first-person scenarios in egocentric video understanding \cite{zhang2022actionformer, lin2022egocentric}. By leveraging content from real-world YouTube videos through an automatic pipeline, YT-Conversation aims to provide a more scalable resource for turn-taking pretraining.

\begin{figure}[t]
\centering
\includegraphics[trim={0 0cm 0 0},width=1.0\linewidth]{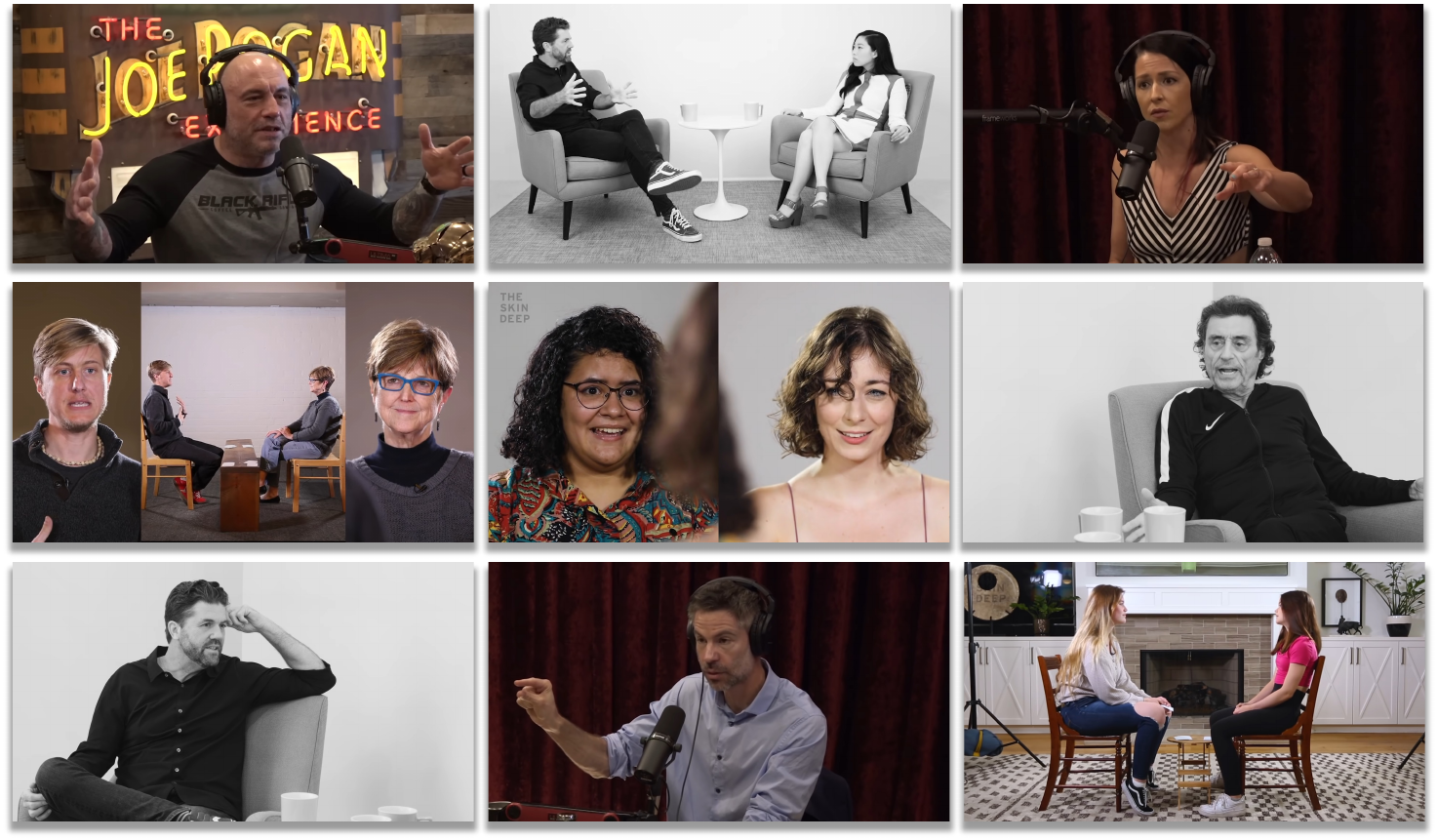}
\caption{Sample frames from YT-Conversation dataset. The dataset includes a diverse range of conversational scenarios from YouTube, such as podcasts, interviews, and informal dialogues, representing various real-world conversation formats.}
\vspace*{-1em}
\end{figure}

\paragraph{Collecting Conversational Videos.}

We curated our dataset from four manually selected YouTube channels, covering diverse conversational formats including podcasts, interviews, and face-to-face dialogues. Videos were randomly sampled without further filtering, ensuring scalability.  
We preprocessed the videos by downsampling to 20 FPS for video and 16 kHz for audio, and trimmed opening segments. Our final dataset comprises 414 videos totaling 41 hours, with durations ranging from 1 to 60 minutes.

\paragraph{Pseudo Per-frame Annotation for Collected Videos}

\begin{figure}[t]
\centering
\includegraphics[trim={0 0cm 0 0},width=1\linewidth]{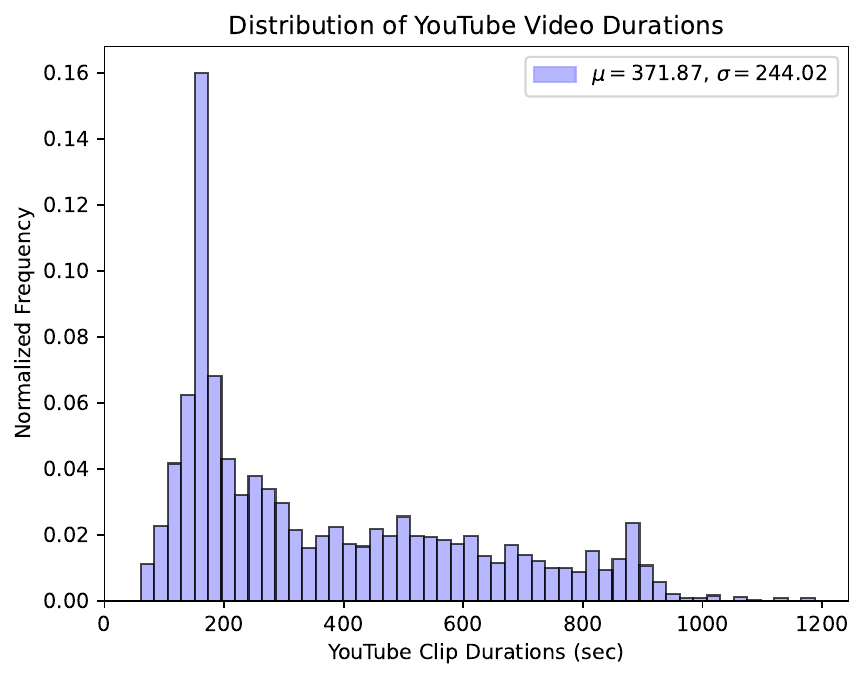}
\caption{Video duration distribution for YT-Conversation. Our online formulation allows the use of long video clips, some even exceeding 900 seconds.}
\label{fig:fig5_YT_video_dist}
\vspace*{-1.0em}
\end{figure}

Since manual annotation of each video frame is labor-intensive, we employ voice activity detection (VAD) from Pyannote \cite{Plaquet23powerset, Bredin23pyanote} to generate pseudo-labels for speech activity. Specifically, we remove any speech segments under 200\,ms 
(or non-speech gaps under 200\,ms) to match our 200\,ms resolution. This approach yields 
a speech vs.\ no-speech label per frame, effectively approximating the ground truth 
for large-scale pretraining. \Cref{fig:fig5_YT_video_dist} shows a distribution of 
video durations and illustrates the diversity of conversation styles in YT-Conversation. For the validation of pseudo-annotation quality, see \Cref{app:YTConv_quality_validation}.

\section{Experimental Setup}\label{sec:experiments}
\subsection{Dataset}

We propose to use publicly available egocentric conversational video datasets for evaluation: EasyCom \cite{donley2021easycom} and Ego4D \cite{grauman2022ego4d}.

\paragraph{EasyCom.}
The EasyCom dataset contains egocentric videos of 3-5 participants conversing around a table in a room for about 30 minutes per session. It comprises 12 sessions totaling approximately 5 hours and 18 minutes. We use sessions 1-3 for testing and 4-12 for training. The dataset features human-annotated transcripts with precise timestamps and mono-channel audio.

\paragraph{Ego4D.}
We use the Audio-Visual Diarization benchmark from Ego4D \cite{grauman2022ego4d}, a large-scale, in-the-wild egocentric video dataset. This subset contains 5-minute clips from diverse scenarios, including both indoor and outdoor settings. However, the original test split is mostly limited to indoor settings. To ensure robust evaluation, we randomly split the combined original train and test sets into 346 training clips and 87 test clips.

The EasyCom and Ego4D datasets offer complementary scenarios for evaluating our framework. EasyCom provides a controlled setting with continuous conversations among fixed participants, while Ego4D presents diverse, real-world scenarios with varying numbers of speakers, environments, and intermittent speech patterns. This combination allows us to assess EgoSpeak's performance in both relatively structured and unstructured environments, testing its ability to predict utterance initiation across a range of conversational dynamics.

\begin{table*}[t]
\centering
\footnotesize

\begin{subtable}{\textwidth}
\centering
\begin{adjustbox}{width=\textwidth, center}
\begin{tabular}{llccccccccccc}
\toprule
\multirow{2}{*}{Model} & \multirow{2}{*}{Modality} & \multicolumn{11}{c}{mAP (\%)} \\
\cmidrule(lr){3-13}
 & & 0.20s & 0.40s & 0.60s & 0.80s & 1.00s & 1.20s & 1.40s & 1.60s & 1.80s & 2.00s & Avg \\
\midrule
\multirow{4}{*}{\centering Transformer} 
 & A & 72.2 & 65.2 & 60.3 & 56.8 & 54.4 & 53.1 & 52.4 & 52.0 & 51.6 & 51.4  & 56.9 $\pm$ 0.05 
 \\
 & V & 52.0 & 51.7 & 51.6 & 51.3 & 51.1 & 50.9 & 50.8 & 50.5 & 50.3 & 50.1  & 51.0 $\pm$ 0.08 
 \\
 & A+V & 73.8 & 66.9 & 62.1 & 58.5 & 56.3 & 55.0 & 54.1 & 53.7 & 53.3 & 53.0 & 58.7 $\pm$ 0.13
 \\
 & A+V\textsuperscript{P} & 73.4 & 66.8 & 61.8 & 58.3 & 56.1 & 54.8 & 54.1 & 53.5 & 53.2 & 52.7  & 58.5 $\pm$ 0.26 
 \\
\midrule
\multirow{4}{*}{\centering GRU} 
 & A & 71.5 & 65.0 & 60.1 & 57.0 & 55.0 & 53.8 & 52.9 & 52.2 & 51.5 & 50.9  & 57.0 $\pm$ 0.30 
 \\
 & V & 53.0 & 52.7 & 52.4 & 52.0 & 51.7 & 51.6 & 51.2 & 51.1 & 50.8 & 50.6 & 51.7 $\pm$ 0.29 
 \\
 & A+V & 73.5 & 68.1 & 63.7 & 60.7 & 59.1 & 58.1 & 57.2 & 56.3 & 55.4 & 54.4 & 60.6 $\pm$ 0.17 
 \\
 & A+V\textsuperscript{P} & 70.8 & 64.9 & 60.1 & 56.9 & 55.0 & 53.8 & 53.0 & 52.4 & 51.8 & 51.4 & 57.0 $\pm$ 0.29 
 \\
\midrule
\multirow{4}{*}{\centering Mamba} 
 & A & 67.5 & 62.2 & 58.4 & 55.7 & 54.0 & 52.9 & 52.0 & 51.1 & 50.2 & 49.6 & 55.4 $\pm$ 0.62 
 \\
 & V & 52.2 & 51.8 & 51.5 & 51.1 & 50.9 & 50.7 & 50.5 & 50.4 & 50.0 & 49.7 & 50.9 $\pm$ 0.21 
 \\
 & A+V & 71.8 & 65.4 & 60.5 & 57.1 & 55.0 & 53.9 & 53.5 & 53.1 & 52.3 & 51.8 & 57.4 $\pm$ 0.26 
 \\
 & A+V\textsuperscript{P} & 68.9 & 63.2 & 59.1 & 56.0 & 54.0 & 52.7 & 51.8 & 51.4 & 50.7 & 50.1 & 55.8 $\pm$ 0.43 
 \\
\bottomrule
\end{tabular}
\end{adjustbox}
\label{tab:main_result_a}
\caption*{(a) Results on EasyCom} %
\end{subtable}

\begin{subtable}{\textwidth}
\centering
\begin{adjustbox}{width=\textwidth, center}
\begin{tabular}{llccccccccccc}
\toprule
\multirow{2}{*}{Model} & \multirow{2}{*}{Modality} & \multicolumn{11}{c}{mAP (\%)} \\
\cmidrule(lr){3-13}
 & & 0.20s & 0.40s & 0.60s & 0.80s & 1.00s & 1.20s & 1.40s & 1.60s & 1.80s & 2.00s & Avg \\
\midrule
\multirow{4}{*}{\centering Transformer} 
 & A & 78.8 & 74.9 & 71.8 & 69.7 & 68.1 & 67.0 & 66.3 & 65.7 & 65.1 & 64.7 & 69.2 $\pm$ 0.03
 \\
 & V & 58.7 & 58.5 & 58.4 & 58.2 & 58.1 & 58.0 & 57.9 & 57.8 & 57.7 & 57.7 & 58.0 $\pm$ 0.27
 \\
 & A+V & 78.1 & 74.3 & 71.5 & 69.4 & 68.0 & 67.0 & 66.3 & 65.7 & 65.3 & 64.9 & 69.0 $\pm$ 0.24
 \\
 & A+V\textsuperscript{P} & 78.4 & 74.5 & 71.5 & 69.4 & 67.9 & 66.7 & 65.9 & 65.4 & 65.0 & 64.5 & 68.9 $\pm$ 0.18
\\
\midrule
\multirow{4}{*}{\centering GRU} 
 & A & 78.6 & 74.8 & 71.8 & 69.6 & 68.1 & 66.9 & 66.2 & 65.6 & 65.2 & 64.8  & 69.2 $\pm$ 0.25
 \\
 & V & 58.6 & 58.3 & 58.1 & 57.9 & 57.8 & 57.8 & 57.7 & 57.6 & 57.5 & 57.5  & 57.9 $\pm$ 0.61
 \\
 & A+V & 76.4 & 73.0 & 70.4 & 68.5 & 67.1 & 66.3 & 65.6 & 65.2 & 64.7 & 64.4  & 68.2 $\pm$ 0.42
 \\
 & A+V\textsuperscript{P} & 76.9 & 73.4 & 70.6 & 68.6 & 67.3 & 66.3 & 65.6 & 65.1 & 64.7 & 64.4  & 68.3 $\pm$ 0.18
 \\
\midrule
\multirow{4}{*}{\centering Mamba} 
 & A & 77.4 & 73.6 & 70.5 & 68.5 & 66.9 & 65.8 & 65.0 & 64.3 & 63.9 & 63.5  & 67.9 $\pm$ 0.37
 \\
 & V & 58.2 & 58.1 & 57.9 & 57.8 & 57.6 & 57.5 & 57.5 & 57.4 & 57.4 & 57.3 & 57.7 $\pm$ 0.28
 \\
 & A+V & 76.0 & 72.5 & 69.8 & 67.9 & 66.6 & 65.6 & 64.8 & 64.2 & 63.8 & 63.5  & 67.5 $\pm$ 0.18
 \\
 & A+V\textsuperscript{P} & 74.1 & 70.8 & 68.1 & 66.2 & 64.8 & 63.9 & 63.2 & 62.7 & 62.3 & 62.0 & 65.8 $\pm$ 0.23
 \\
\bottomrule
\end{tabular}
\end{adjustbox}
\label{tab:main_result_b}
\caption*{(b) Results on Ego4D} 
\end{subtable}

\caption{Mean average precision (mAP) scores on (a)~EasyCom and (b)~Ego4D at time steps 0.20\,s to 2.00\,s for Transformer, GRU, and Mamba architectures under Audio~(A), Visual~(V), and Audio+Visual~(A+V) modalities. Models with \textsuperscript{P} are pretrained on YT-Conversation. Per-timestep values come from a single random seed, while ``Avg'' shows mean $\pm$ SE over five seeds (see \Cref{app:stat_results} for full multi-seed results). Using both A and V yields the best performance overall.}

\label{tab:main_result}
\end{table*}

\begin{table}[t]
\footnotesize
\centering
\begin{adjustbox}{center}
\begin{tabular}{llc}
    \noalign{\hrule height 1pt}
    &&\\[-2ex]
    Dataset  &   Model   &  Target Speaker AP   \\
    &&\\[-2.5ex]
    \hline
    &&\\[-2ex]
    \multirow{3}{*}{EasyCom}
                            &   Transformer (A+V) &  \textbf{52.7}     \\
                            &   Random       &  27.2     \\
                            &   Silence-based       &  26.6     \\

    &&\\[-2.5ex]
    \hline
    &&\\[-2ex]
    \multirow{3}{*}{Ego4D}   & Transformer (A+V) &  \textbf{66.8}    \\
                              &  Random     &   26.1      \\
                              &  Silence-based     &   27.7      \\
                              
    &&\\[-2.5ex]
    \noalign{\hrule height 1pt}
\end{tabular}
\end{adjustbox}
\caption{Performance comparison between our predictive Transformer model (A+V) and detection-based baselines on EasyCom and Ego4D datasets. }
\vspace*{-1em}
\label{tab:avg_perframe_ap}
\end{table}

\subsection{Baselines \& Models}
We evaluate our framework using three trained models with different architectural backbones: RNN~\cite{an2023miniroad}, Transformer~\cite{xu2021longlstr}, and State-Space-Model~\cite{gu2023mamba}. Additionally, we implement two static baselines: a random baseline and a rule-based algorithm using silence as decision threshold~\cite{bell2001real}. Detailed implementation details including architectural specifications, hyperparameters, training objective and feature extraction for the neural models are provided in \Cref{app:implementation}.

\paragraph{Random Baseline.}
This baseline randomly assigns one of the three possible labels (background, target speaker speaking, or other speaking) to each frame with uniform probability.

\paragraph{Silence-based Algorithm.}
Simulating commercial spoken dialogue agents, this approach triggers speech only after a 600\,ms silence interval following other speakers. Our evaluation likely overestimates its real-world performance, since we use ground-truth labels to detect non-target speech and count the entire subsequent speech segment as correct once the start is identified (only a single timestep is penalized when incorrect).

\paragraph{Transformer-based Model.}
We adopt Long Short-term TRansformer (LSTR) \cite{xu2021longlstr} for temporal modeling. LSTR uses long-term and short-term memory mechanisms to handle sequence data, with an encoder-decoder structure. The encoder leverages long context windows by compressing inputs, while the decoder processes shorter context windows, allowing for flexible temporal modeling.

\paragraph{RNN-based Model.}
Inspired by \citet{an2023miniroad}, we used a simple and effective RNN model containing one GRU layer. This model was chosen for its computational efficiency and strong performance. 

\paragraph{Mamba-based Model.}
We implement a Mamba-based model \cite{gu2023mamba} similar to the RNN architecture. Given the recent success of Mamba across various tasks, we include this model to explore its potential to predict speech initiation in egocentric videos while maintaining computational efficiency.

\subsection{Settings}
\paragraph{Feature Extraction.} \label{para:feature_extraction}
We process features at 5 FPS, predicting every 0.2 seconds to align with typical human response times \cite{skantze2021turnreview}. For RGB features, we use a ResNet-50 \cite{he2016resnet} model pretrained on Kinetics-400 \cite{kay2017kinetics}. Audio features are extracted using wav2vec2 \cite{baevski2020wav2vec}. These features are concatenated to create our multimodal input. Further details on feature extraction are provided in \Cref{app:feature_extraction}.

\paragraph{Evaluation Protocol.}

Most existing turn-taking evaluations rely on offline F1-scores after processing the entire clip \cite{lee2023multimodal, kurata2023multimodal}, or on sample-based F1-scores around turn-taking events using threshold-based detection \cite{ekstedt2022voice, onishi2023multimodalvap}. However, both approaches fail to capture the continuous, overlapping nature of real-world conversations, where a decision must be made at every frame. As \citet{heldner2010pauses} suggested, overlaps occur frequently in human conversation. Consequently, we measure performance per frame to better reflect these natural conversational dynamics.

To address this need, we propose using per-frame mean average precision (mAP), inspired by 
prior work on online tasks \cite{de2016online}. This metric evaluates how well the model 
anticipates the target speaker’s speech up to 10 timesteps (2\,s) into the future. We compute mAP by 1) sorting all frame-level confidence scores in descending order, 2) iteratively using each score as a threshold, 3) calculating precision and recall at each threshold, and 4) averaging all precision values. This procedure is repeated for each class and timestep, then averaged to yield the final mAP.

\section{Results and Analysis}\label{sec:results}
\subsection{Quantitative Results}\label{subsec:quant}
\Cref{tab:main_result,tab:avg_perframe_ap} present comprehensive comparisons of our models across different modalities, datasets, and baselines. To ensure the robustness of our findings, we report performance over five random seeds with error bars in \Cref{app:stat_results}.
\paragraph{Model Performance Across Modalities}
As shown in \Cref{tab:main_result}, the multimodal (A+V) approach generally outperforms unimodal inputs on EasyCom, with the Transformer model achieving 58.7\% mAP (compared to 56.9\% for audio-only and 51.0\% for visual-only). The GRU model performs 
best with A+V (60.6\% mAP), while Mamba sees moderate improvements (57.4\% mAP). On Ego4D, the Transformer with A+V attains 69.0\% mAP, which is roughly on par with its audio-only counterpart. Interestingly, GRU and Mamba actually do better with audio alone, at 69.0\% and 67.9\% mAP respectively.

\paragraph{Pretraining Effects}
Our YT-Conversation pretraining results show that overall gains are modest. However, 
we do observe a small but consistent improvement in detecting \emph{other person speaking} class, which is especially valuable in egocentric scenarios. In contrast, GRU and Mamba show 
little or no net gain, aligning with prior work that certain recurrent/state-space models 
often struggle with large-scale pretraining \cite{wang2023pretraining}. 
We attribute these results to domain mismatch and the inherently noisier nature of real-world conversational videos. For a per-class breakdown and further discussion, refer to \Cref{app:error_analysis}.

\paragraph{Comparison with Baselines}
\Cref{tab:avg_perframe_ap} compares our best Transformer (A+V) model with both baselines 
on EasyCom and Ego4D. Even though the silence-based approach benefits from an evaluation bias, 
our predictive model still achieves significantly higher AP (52.7\% vs.\ 26.6\% on EasyCom, and 66.8\% vs.\ 27.7\% on Ego4D). Moreover, the silence-based method performs similarly to random, indicating that requiring a fixed silence interval fails to accommodate the fluid, overlapping speech found in real-world conversations.

\begin{table}
\footnotesize
\centering
\begin{adjustbox}{center}
\begin{tabular}{llc}
    \noalign{\hrule height 1pt}
    &&\\[-2ex]
    Model  &   Modality   &   Avg. mAP   \\
    &&\\[-2.5ex]
    \hline
    &&\\[-2ex]
    \multirow{3}{*}{Transformer}   &   A (+F)       &  57.0 (+8.0)     \\
                            &   V (+F)       & 51.2 (+6.4)      \\
                            &   A+V (+F)     &  \textbf{58.8 (+9.6)}     \\
    &&\\[-2.5ex]
    \hline
    &&\\[-2ex]
    \multirow{3}{*}{GRU}   &   A (+ F)      &   57.1 (+3.9)    \\
                                &   V (+ F)      &   51.2 (+4.2)    \\
                                &   A+V (+ F)    &   61.2 (+0.9)    \\
    &&\\[-2.5ex]
    \hline
    &&\\[-2ex]
    \multirow{3}{*}{Mamba} &   A (+ F)      &   55.6 (+2.6)   \\
                                &   V (+ F)      &   51.3 (+3.9)   \\
                                &   A+V (+ F)    &   57.6 (+2.7)   \\
    \noalign{\hrule height 1pt}

\end{tabular}
\end{adjustbox}
\caption{Impact of optical flow on utterance initiation prediction for EasyCom dataset. Values show average mAP, with performance gains from optical flow in parentheses. A: Audio, V: Visual, F: Flow. Bold indicates the best overall performance.}
\label{tab:ablation_optical_flow}
\end{table}

\begin{figure}[t]
\centering
\includegraphics[width=1.0\linewidth]{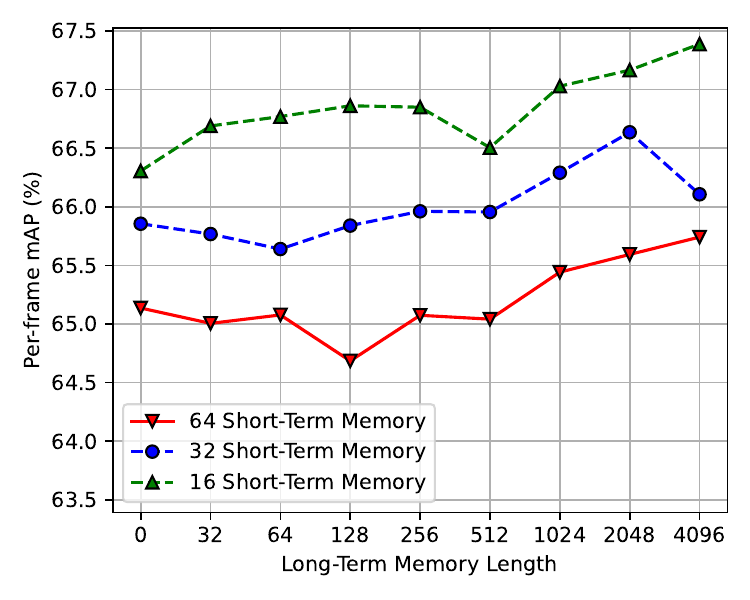}
\caption{Utterance initiation prediction with varying transformer memory length. a shorter context window for short-term memory and a longer context window for long-term memory generally show better results.}
\label{fig:context_length}
\end{figure}

\begin{figure*}[t]
\centering
\includegraphics[width=1\linewidth]{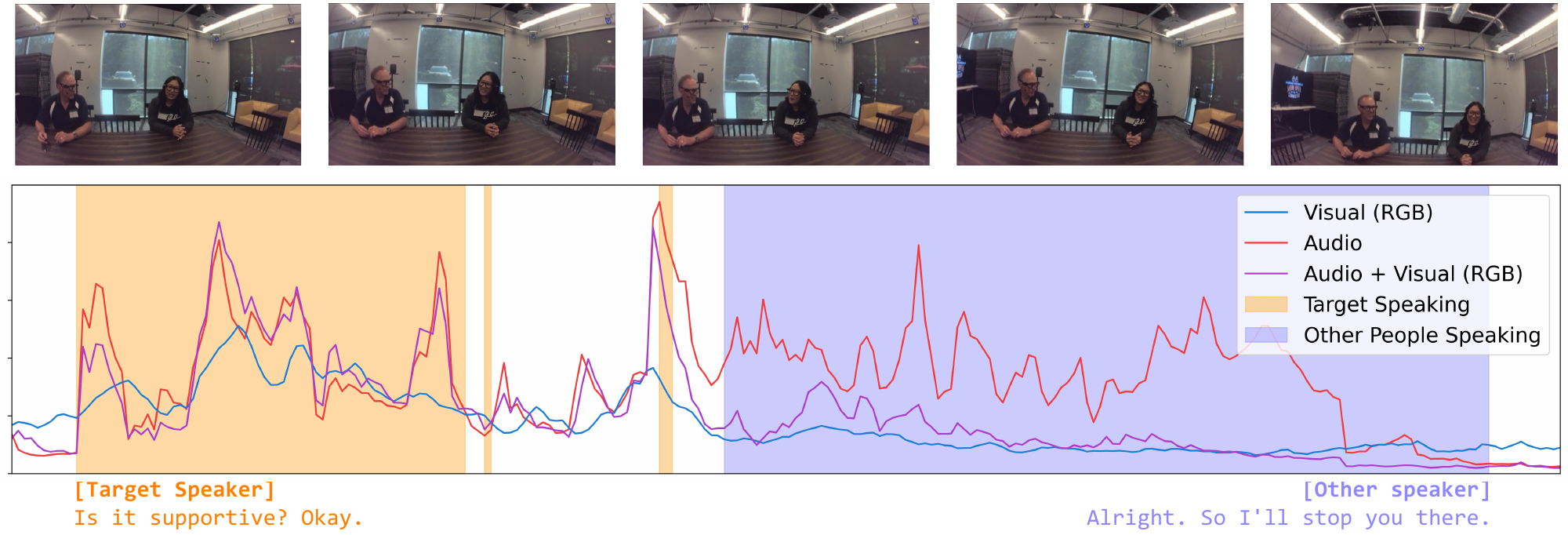}
\caption{Qualitative results on EasyCom. The predicted scores are shown in lines and the ground-truth label is shown in regions. The blue line represents a model with RGB input, the red line represents a model with audio input, and the purple line represents a model with audio and visual input.}
\label{fig:qualitative_results}
\end{figure*}

\subsection {Motion Inputs Contribute to Turn-Taking Prediction}
Since many non-verbal cues involve motion, we hypothesized that incorporating optical flow could improve utterance initiation prediction.
To test this, we extracted optical flow using the Denseflow toolkit \cite{wang2020denseflow} with the TV-L1 algorithm \cite{zach2007dualityTVL1}, following a similar process to our RGB feature extraction. Optical flow is a computer vision technique that estimates object motion between consecutive video frames by calculating the apparent motion of brightness patterns. This is useful for tracking movement and analyzing dynamic scenes.

\Cref{tab:ablation_optical_flow} presents the results of our experiment on EasyCom. Incorporating optical flow consistently improved performance across all model types and input combinations.
These results suggest that motion information provides valuable cues for predicting utterance initiation, complementing static visual and audio features to enable more accurate predictions of speech onset.

\subsection{Models Do Not Exploit Short-Term Information Well}
Our framework uses online processing models that rely on context length to capture historical information. Because dialogue context is crucial in turn-taking \cite{skantze2021turnreview}, 
the choice of context can strongly affect utterance initiation. \Cref{fig:context_length} shows how the Transformer model’s performance varies with different long-term and short-term window sizes. While extending the long-term window helps, increasing the short-term window unexpectedly degrades performance. This suggests that although a broader context provides valuable cues, an overly large short-term window may introduce noise or irrelevant data, reducing accuracy. These findings highlight the importance of balancing long-term and short-term context in untrimmed videos to optimize turn-taking predictions.

\subsection{Runtime Analysis}
We evaluated the computational efficiency of our models by measuring their frames per second (FPS), parameter counts, and floating-point operations (GFLOPs) on a single RTX3090 GPU using the EasyCom dataset, as shown in \Cref{tab:fps}. The RNN-based model achieved the highest throughput with 13,939.5 FPS, while maintaining a relatively low parameter count of 34.6M and requiring 206.52 GFLOPs. The Mamba-based model followed with 12,009.3 FPS, though it has the highest parameter count (83.1M) and the largest computational requirement (610.93 GFLOPs). The Transformer-based model ran at 99.8 FPS with 67.21M parameters and 129.48 GFLOPs, achieving real-time processing but at a lower frame rate than RNN and Mamba. Overall, these results indicate that all three architectures are capable of real-time processing.

\begin{table}[t]
\footnotesize
\centering
\begin{adjustbox}{width=1\columnwidth}
\begin{tabular}{lccc}
\toprule
Model & FPS & Parameters & GFLOPs \\
\midrule
Transformer & 99.8 & 67.21M & 129.48 \\
RNN & 13939.5 & 34.6M & 206.52 \\
Mamba & 12009.3 & 83.1M & 610.93 \\
\bottomrule
\end{tabular}
\end{adjustbox}
\caption{Runtime analysis of different models on a single RTX3090 GPU.}
\vspace*{-0.5em} 
\label{tab:fps}
\end{table}

\subsection {Qualitative Results}
\Cref{fig:qualitative_results} shows the qualitative results based on Transformer. Our observations indicate that the model using only RGB features struggles to effectively distinguish between speaking and non-speaking segments, leading to frequent misclassifications. In contrast, the model utilizing audio input shows notable improvement in predicting the target speaker's speech. However, the audio-only model often assigns high probabilities to the speech of other individuals, resulting in less accurate turn-taking. Notably, the model that integrates both audio and visual inputs demonstrates superior performance. This multimodal model effectively distinguishes the target speaker from others, accurately identifying speaking segments while minimizing false positives from other speakers.

\section{Conclusion}\label{sec:conclusion}
We introduced \frameworkname, a novel framework for real-time speech initiation prediction from an in-the-wild, first-person viewpoint. \frameworkname integrates four key capabilities to better handle complex, dynamic real-world conversations. We also presented YT-Conversation, a large-scale dataset of in-the-wild YouTube videos for pretraining.

Empirical results on two egocentric conversational video datasets demonstrate that \frameworkname significantly outperforms random and silence-based baselines in real time. Notably, our analysis revealed that incorporating optical flow significantly improves performance and highlighted a counterintuitive finding on context length. We encourage further exploration of large-scale conversational datasets, improved multimodal modeling techniques, and integration with large language models to generate responses, advancing turn-taking in real-world settings.

\section{Limitations}\label{sec:limitations}
Our proposed method relies on pre-encoded features, which can limit both performance and frames per second (FPS). A potential solution is to adopt an end-to-end framework that learns to extract relevant features directly from raw input. For example, E2E-LOAD \cite{cao2023e2e} demonstrates improvements in both state-of-the-art performance and FPS in online action recognition task through such an approach.

While our YT-Conversation dataset provides diverse pretraining data, it may not fully capture the nuances of first-person interactions. Future work could explore methods to augment the dataset with more egocentric conversational data, potentially improving model performance in first-person scenarios.

Lastly, our current approach does not explicitly model speaker-specific behaviors. Future research could incorporate individual speaking patterns and tendencies, by analyzing larger egocentric conversational datasets. By capturing these speaker-specific nuances, future models may better anticipate utterance initiation points, particularly in prolonged conversations with familiar participants.

\section{Ethical Statement}\label{sec:Ethical Statement}

\paragraph{Data Collection and Privacy Considerations}
Although the YT-Conversation, derived from publicly shared YouTube videos, provides natural conversations for training AI models, there is a possibility it may capture the facial features of the participants. However, the YT-Conversation dataset was collected under the principles of informed consent and data anonymization, adhering to the ACM Code of Ethics 1.6 (Respect privacy).

\paragraph{Informed consent} We selected videos that participants are likely aware of and have consented to be recorded and publicly shared, such as podcasts, interviews, and face-to-face conversations.
    
\paragraph{Data Anonymization} We only released the YouTube IDs rather than the raw YouTube videos so that content creators can remove their videos from YouTube anytime, which will automatically exclude them from our dataset. Moreover, our transcripts only include the time ranges for the start and end of the speech, along with the corresponding video frames without personal information.

\section{Acknowledgements}
\label{sec:acknowledgement}

We thank Hyolim Kang and Joungbin An for their valuable discussions and insightful feedback.
This work was supported by Artificial intelligence industrial convergence cluster development project funded by the Ministry of Science and ICT(MSIT, Korea)\&Gwangju Metropolitan City and NCSOFT. It was also partly supported by an IITP grant funded by the Korean Government (MSIT) (No.RS-2020-II201361, Artificial Intelligence Graduate School Program (Yonsei University) and RS-2024-00353131) and the National Research Foundation of Korea (NRF) grant funded by the Korea government (MSIT) (No. RS-2024-00354218).

\bibliography{custom}
% \bibliography{anthology,custom}
% \bibliographystyle{acl_natbib}

\clearpage
\appendix
\section{Implementation Details}
\label{app:implementation}
\subsection{Architecture \& Hyperparameters}
\label{app:arch_hyper}
This section provides detailed information on the architectures and hyperparameters used for each model in our experiments. We set the anticipation length to 10 timesteps for all models, predicting up to 2 seconds into the future. All experiments were done with a single RTX3090 GPU within one day.

\paragraph{Transformer-based Model (LSTR)}
For the transformer model, we configured 16 attention heads and 1024-dimensional hidden units in the transformer blocks. The LSTR encoder processes long context windows up to 2048 frames, while the decoder handles shorter context windows up to 32 frames. We trained this model using the Adam optimizer \cite{kingma2014adam} with a weight decay of $5 \times 10^{-5}$. The learning rate was scheduled to increase linearly from zero to $7 \times 10^{-5}$ during the first 40\% of training iterations, then decrease to zero following a cosine function. We trained the transformer model for 50 epochs with a batch size of 16.

\paragraph{RNN-based Model}
For the RNN model, we used 2048-dimensional embeddings and 1024-dimensional hidden units. This model was trained for 30 epochs with a batch size of 64. We used the same optimizer and learning rate schedule as the transformer model.

\paragraph{Mamba-based Model}
The Mamba-based model builds upon the RNN architecture, replacing the GRU layer with a Mamba block. We set the SSM state factor to 16, local convolution width to 4, and block expansion factor to 2. The training settings were kept consistent with the RNN model.

\subsection{Training Objective}
For training, we use cross-entropy loss between predicted confidence scores $s_T$ at time $T$ and the ground-truth label $y_T \in \{0, 1, \ldots, K\}$. $K$ is the number of classes and $s^k_T$ is the $k$-th element of the probability vector $s_T$. For Transformer-based model, $\alpha_T$ is always 1. For RNN-based and Mamba-based models, $\alpha_T$ is used to modulate the contribution of intermediate time steps during the computation of the loss. Specifically, $\alpha_T$ takes the value 1 only at a designated step $t = L$ and 0 otherwise.

We also define a temporal window of length $L$, which determines the final step contributing to the objective function:

$$J(y_T, s_T; T) = - \sum_{k=0}^K \alpha_T\delta(k - y_T) \log s_T^k,$$

\subsection{Feature Extraction}
\label{app:feature_extraction}
\paragraph{RGB Features} As mentioned in \Cref{para:feature_extraction}, videos are downsampled to 20 FPS and processed in 4-frame chunks, resulting in a 5 FPS prediction rate. We use ResNet-50 \cite{he2016resnet} initialized with weights from a video action recognition model \cite{wang2016temporal}, implemented via MMAction2 \cite{2020mmaction2}. The center frame of each chunk is sampled for feature extraction. For the EasyCom dataset, we cropped all clips in each session to remain only video frames and merged them to make one video per session.
\paragraph{Audio Features} We use wav2vec2's \cite{baevski2020wav2vec} multi-layer convolutional feature encoder, as noted in \Cref{para:feature_extraction}. Every 10 encoded audio features are concatenated temporally to match the 5 FPS RGB features.

\begin{figure}[t]
\centering
\includegraphics[width=0.99\linewidth]{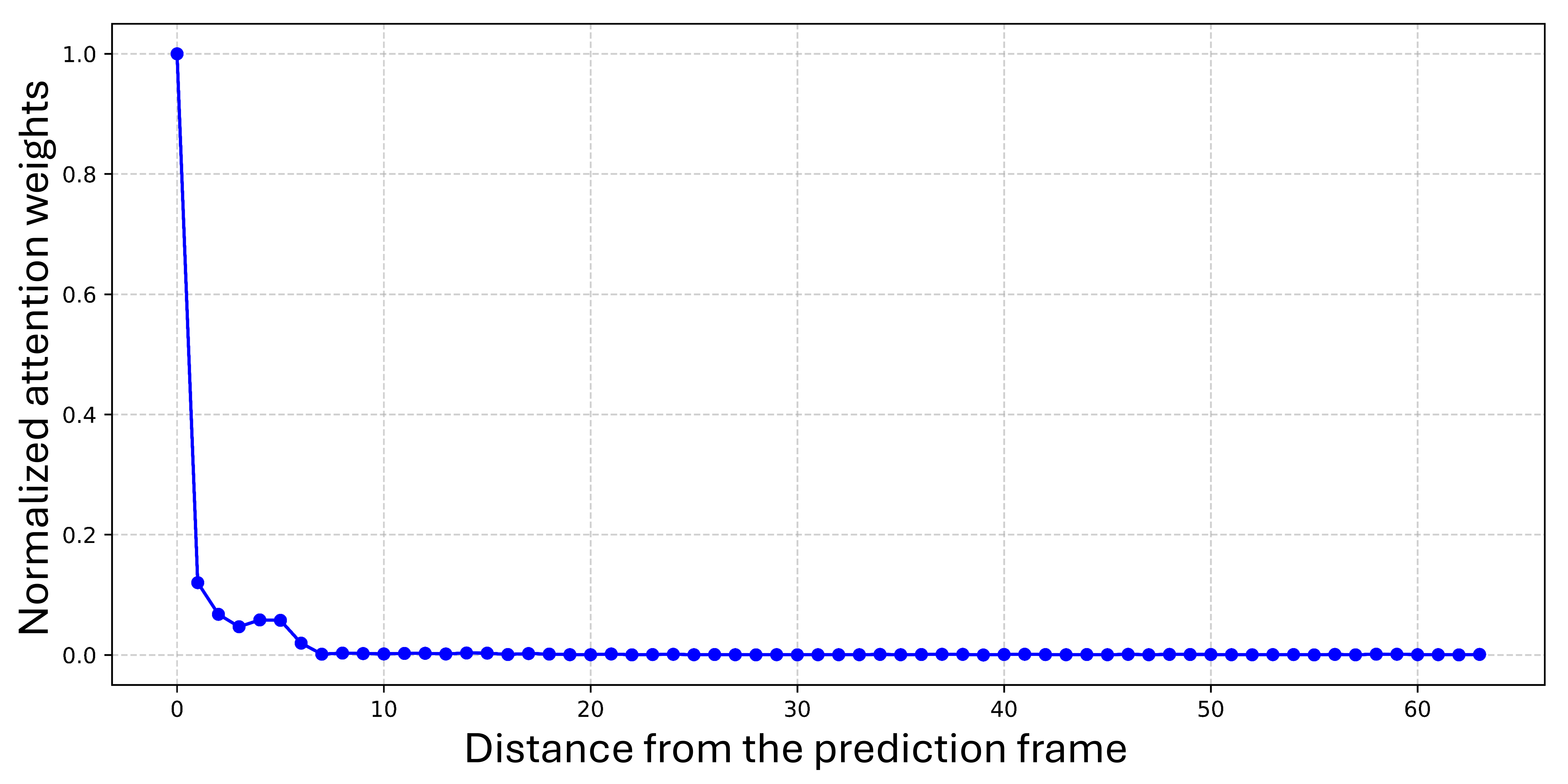}
\caption{Attention weight of a transformer encoders. Transformer models focus on mostly local context for utterance initiation.}
\label{fig:attn_heatmap}
\vspace*{-1.5em}
\end{figure}

\section {Importance of Recent Frames}

\Cref{fig:attn_heatmap} shows the distribution of attention weights across the encoder layers of a transformer model in the context of predicting utterance initiation in real-world conversations \cite{wang2021oadtr}. The attention weights of the test set were averaged with respect to the layers, multi-heads, and batch and then normalized.  These weights reveal the significance assigned to each frame in the sequence during prediction. Our analysis shows that the model focuses predominantly on the recent frames, with attention weights diminishing notably as the distance from the current frame increases. This pattern indicates that recent frames have a greater impact on the model's predictions for utterance initiation. 

\begin{table}[t]
\centering
\footnotesize
\begin{adjustbox}{width=\columnwidth, center}
\begin{tabular}{llccccc}
\toprule
\multirow{2}{*}{Dataset} & \multirow{2}{*}{Model} &
\multirow{2}{*}{\shortstack{Avg\\mAP}} & \multicolumn{3}{c}{Per-Class AP (\%)} \\
\cmidrule(lr){4-6}
& & & Background & Target Speaker & Other Speaker \\
\midrule
\multirow{2}{*}{EasyCom} 
  & Transformer & 58.79 & 43.17 & 52.74 & 80.46 \\
  & Transformer\textsuperscript{P} & 59.01 & 42.94 & 52.91 & 81.17 \\
\midrule
\multirow{2}{*}{Ego4D} 
  & Transformer & 69.61 & 73.50 & 66.78 & 68.56 \\
  & Transformer\textsuperscript{P} & 68.79 & 71.80 & 64.46 & 70.11 \\
\bottomrule
\end{tabular}
\end{adjustbox}
\caption{Per-class average precision (AP) for the Transformer with and without 
(\textsuperscript{P}) YT-Conversation pretraining on EasyCom and Ego4D. Although overall 
gains are modest, we observe a notable improvement for \emph{Other Speaker} detection.}
\label{tab:pretrain_perclass}
\end{table}

\begin{figure} %
\centering
\includegraphics[trim={0 0cm 0 0},width=1\linewidth]{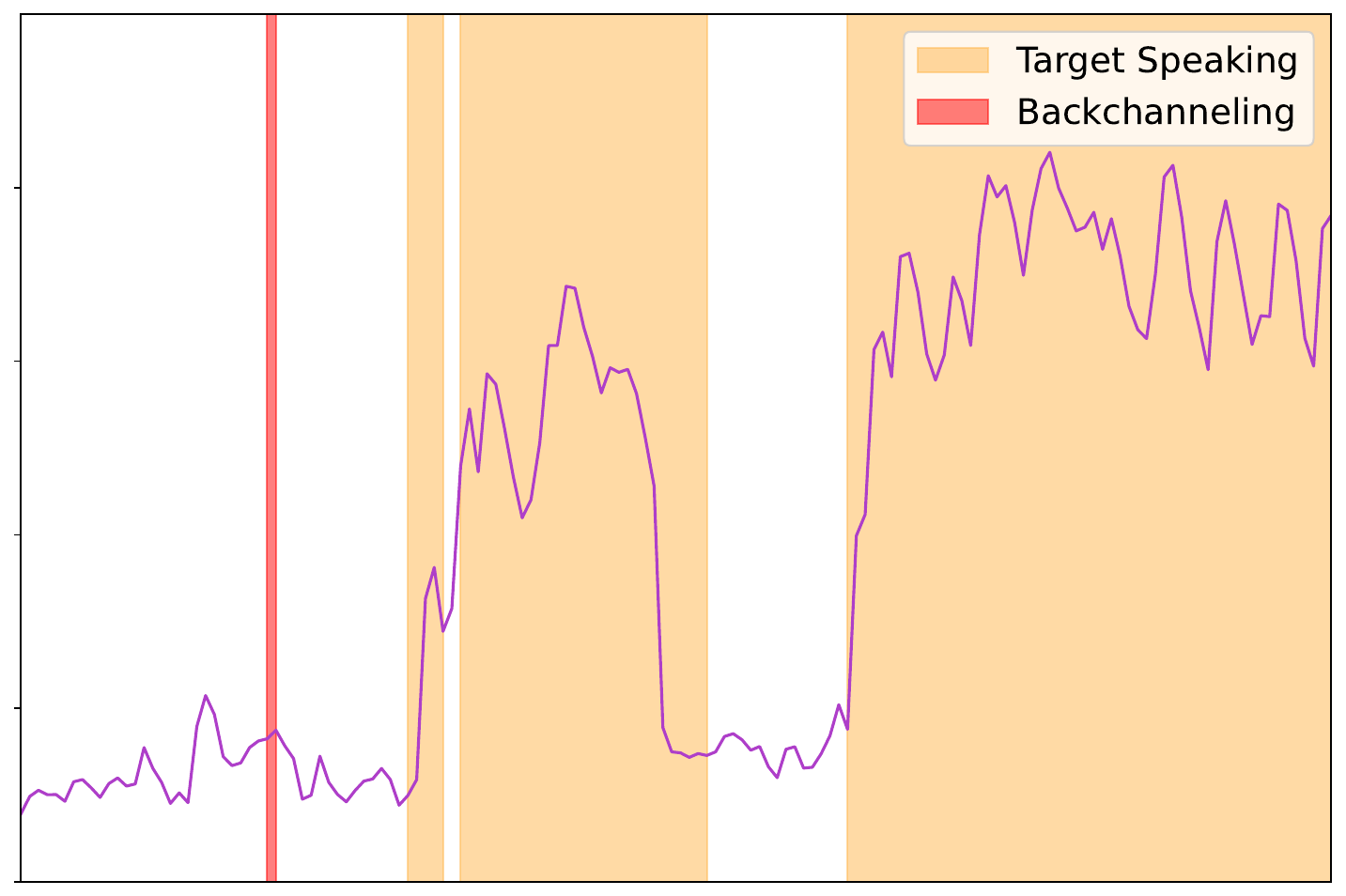}
\caption{Failure case on EasyCom. The orange region represents the target speaker speaking, and the red region represents the target speaker's backchanneling. }
\label{fig:fig9_error_analysis}
\end{figure}

\section{Error Analysis}
\label{app:error_analysis}
\paragraph{Pretraining on YT-Conversation}

We observed that YT-Conversation pretraining yields modest overall gains, but a notable improvement for \emph{other person speaking} class (+0.7\% on EasyCom, +1.5\% on Ego4D). 
\Cref{tab:pretrain_perclass} lists the per-class average precision (AP) for the Transformer 
model with and without pretraining. Although this benefit can be crucial in egocentric 
scenarios—where identifying others’ speech fosters smoother turn-taking—gains for 
\textit{Background} and \textit{Target Speaker} remain unchanged or slightly negative. 
We attribute this to domain mismatch (YouTube interviews vs.\ dynamic ego footage) 
and noisy data from real-world conversational videos such as visual effects or subtitles. Future work might address these limitations by bridging domain gaps—e.g., with domain adaptation—or introducing video filtering to obtain higher-quality conversational clips.

\paragraph{Backchannels}
While our method aims to predict any utterance initiation point, there is a short and brief response that occurs when one participant is speaking and the listener reacts to signify the listener's attention, understanding, or emotion rather than take turns and speak. This behavior is referred to as ``backchannels'' \cite{yngve1970getting, skantze2021turnreview}. We observed that prediction scores usually do not increase before backchanneling. \Cref{fig:fig9_error_analysis} illustrates this phenomenon, showing how the model's prediction scores do not significantly increase before a backchanneling event, in contrast to regular speaking turns.

\section{Descriptive Statistics of Experimental Results}
\label{app:stat_results}

We evaluated each model with five random seeds \{0, 10, 20, 29, 42\} to measure performance variance. \Cref{tab:different_seed_results} shows the multi-seed mean average precision (mAP) on EasyCom and Ego4D, while \Cref{tab:main_result_error_bar_easycom,tab:main_result_error_bar_ego4d} provide per-timestep results (mean $\pm$ standard error). These tables complement the main text figures (\Cref{tab:main_result,tab:avg_perframe_ap}) by offering a full breakdown of multi-seed performance at each time step, ensuring transparency and robustness in our results.

\begin{table}[t]
\footnotesize
\centering
\begin{adjustbox}{width=\columnwidth}
\begin{tabular}{llcc}
    \noalign{\hrule height 1pt}
    &&\\[-2ex]
    Model  &   Modality & EasyCom  & Ego4D \\
    &&\\[-2.5ex]
    \hline
    &&\\[-2ex]
    \multirow{4}{*}{Transformer}   &   A      &  56.9 $\pm$ 0.12 & 69.2 $\pm$ 0.08 \\
                            &   V      & 51.0 $\pm$ 0.18 & 58.1 $\pm$ 0.60 \\
                            &   A+V   &  58.7 $\pm$ 0.29 & 69.6 $\pm$ 0.54 \\
                            &   A+V\textsuperscript{P}  &  58.5 $\pm$ 0.59 & 68.8 $\pm$ 0.41 \\
    &&\\[-2.5ex]
    \hline
    &&\\[-2ex]
    \multirow{4}{*}{GRU}   &   A      &   57.0 $\pm$ 0.68 & 69.2 $\pm$ 0.55 \\
                                &   V      &   51.7 $\pm$ 0.65 & 57.9 $\pm$ 1.36 \\
                                &   A+V    &   60.6 $\pm$ 0.38 & 68.2  $\pm$ 0.95 \\
                                &   A+V\textsuperscript{P}   &  57.0 $\pm$ 0.65 & 68.3  $\pm$ 0.41 \\
    &&\\[-2.5ex]
    \hline
    &&\\[-2ex]
    \multirow{4}{*}{Mamba} &   A      &   55.4 $\pm$ 1.39 & 67.9 $\pm$ 0.83\\
                                &   V      &   50.9 $\pm$ 0.48 & 57.7 $\pm$ 0.64 \\
                                &   A+V    &   57.4 $\pm$  0.58 & 67.5 $\pm$ 0.41 \\
                                &   A+V\textsuperscript{P}  &  55.8 $\pm$ 0.97 & 65.8 $\pm$ 0.51   \\
    \noalign{\hrule height 1pt}

\end{tabular}
\end{adjustbox}
\caption{Performance comparison of models across five different seeds on the EasyCom and Ego4D datasets. Each value represents the average mAP across the seeds, along with the standard error.}
\label{tab:different_seed_results}
\end{table}

\section{YT-Conversation Pseudo Annotation Quality Validation}
\label{app:YTConv_quality_validation}
To validate the quality of pseudo-annotations in our YT-Conversation dataset, we conducted a human evaluation study on 100 segments randomly sampled from 10 videos, excluding the first five segments of each (typically non-conversational teasers). Each segment received a label alignment score on a 5-point scale: 
(1) completely misaligned, with timestamps far off from actual speech; 
(2) poor alignment, missing large portions, or labeling silence as speech; 
(3) adequate but potentially off by 0.5–1 second; 
(4) good alignment, within about 0.5 second of true boundaries; and 
(5) excellent alignment, nearly matching human labels. 
Across all evaluated segments, the average alignment score was 2.147. We want to note that as ASR models continue to advance \cite{zusag2024crisperwhisper}, the pseudo-label will be precise as well. We also use these pseudo-labels only for pretraining, ensuring the evaluations remain robust with human-annotated labels.

\begin{table*}[t]
\centering
\footnotesize

\begin{subtable}{\textwidth}
\centering
\begin{adjustbox}{width=\textwidth, center}
\begin{tabular}{llccccc}
\toprule
\multirow{2}{*}{Model} & \multirow{2}{*}{Modality} & \multicolumn{5}{c}{mAP (\%)} \\
\cmidrule(lr){3-7}
 & & 0.20s & 0.40s & 0.60s & 0.80s & 1.00s \\
\midrule
\multirow{4}{*}{\centering Transformer} 
 & A & 72.2 $\pm$ 0.05 & 65.2 $\pm$ 0.08 & 60.3 $\pm$ 0.09 & 56.8 $\pm$ 0.09 & 54.4 $\pm$ 0.08 
 \\
 & V & 52.0 $\pm$ 0.06 & 51.7 $\pm$ 0.08 & 51.6 $\pm$ 0.06 & 51.3 $\pm$ 0.09 & 51.1 $\pm$ 0.06 
 \\
 & A+V & 73.8 $\pm$ 0.12 & 66.9 $\pm$ 0.16 & 62.1 $\pm$ 0.14 & 58.5 $\pm$ 0.16 & 56.3 $\pm$ 0.15 
 \\
 & A+V\textsuperscript{P} & 73.4 $\pm$ 0.17 & 66.8 $\pm$ 0.18 & 61.8 $\pm$ 0.22 & 58.3 $\pm$ 0.31 & 56.1 $\pm$ 0.30
 \\
\midrule
\multirow{4}{*}{\centering GRU} 
 & A & 71.5 $\pm$ 0.30 & 65.0 $\pm$ 0.38 & 60.1 $\pm$ 0.29 & 57.0 $\pm$ 0.30 & 55.0 $\pm$ 0.31   
 \\
 & V & 53.0 $\pm$ 0.28 & 52.7 $\pm$ 0.34 & 52.4 $\pm$ 0.25 & 52.0 $\pm$ 0.34 & 51.7 $\pm$ 0.33  
 \\
 & A+V & 73.5 $\pm$ 0.30 & 68.1 $\pm$ 0.20 & 63.7 $\pm$ 0.30 & 60.7 $\pm$ 0.21 & 59.1 $\pm$ 0.25  
 \\
 & A+V\textsuperscript{P} & 70.8 $\pm$ 0.41 & 64.9 $\pm$ 0.26 & 60.1 $\pm$ 0.29 & 56.9 $\pm$ 0.31 & 55.0 $\pm$ 0.31 
 \\
\midrule
\multirow{4}{*}{\centering Mamba} 
 & A & 67.5 $\pm$ 0.88 & 62.2 $\pm$ 0.98 & 58.4 $\pm$ 0.76 & 55.7 $\pm$ 0.71 & 54.0 $\pm$ 0.61 
 \\
 & V & 52.2 $\pm$ 0.18 & 51.8 $\pm$ 0.18 & 51.5 $\pm$ 0.18 & 51.1 $\pm$ 0.19 & 50.9 $\pm$ 0.17 
 \\
 & A+V & 71.8 $\pm$ 0.22 & 65.4 $\pm$ 0.15 & 60.5 $\pm$ 0.17 & 57.1 $\pm$ 0.14 & 55.0 $\pm$ 0.13 
 \\
 & A+V\textsuperscript{P} & 68.9 $\pm$ 0.51 & 63.2 $\pm$ 0.47 & 59.1 $\pm$ 0.43 & 56.0 $\pm$ 0.44 & 54.0 $\pm$ 0.41 
 \\
\bottomrule
\end{tabular}
\end{adjustbox}
\label{tab:timesteps_different_seeds_EasyCom_a}
\caption*{(a) 5 Different Seeds Results on EasyCom - Timesteps from 0.20s to 1.00s}
\end{subtable}

\begin{subtable}{\textwidth}
\centering
\begin{adjustbox}{width=\textwidth, center}
\begin{tabular}{llccccc}
\toprule
\multirow{2}{*}{Model} & \multirow{2}{*}{Modality} & \multicolumn{5}{c}{mAP (\%)} \\
\cmidrule(lr){3-7}
 & & 1.20s & 1.40s & 1.60s & 1.80s & 2.00s \\
\midrule
\multirow{4}{*}{\centering Transformer} 
 & A &  53.1 $\pm$ 0.06 & 52.4 $\pm$ 0.09 & 52.0 $\pm$ 0.08 & 51.6 $\pm$ 0.07 & 51.4 $\pm$ 0.10 
 \\
 & V &  50.9 $\pm$ 0.06 & 50.8 $\pm$ 0.07 & 50.5 $\pm$ 0.11 & 50.3 $\pm$ 0.12 & 50.1 $\pm$ 0.11 
 \\
 & A+V & 55.0 $\pm$ 0.18 & 54.1 $\pm$ 0.14 & 53.7 $\pm$ 0.21 & 53.3 $\pm$ 0.15 & 53.0 $\pm$ 0.15 
 \\
 & A+V\textsuperscript{P} & 54.8 $\pm$ 0.31 & 54.1 $\pm$ 0.31 & 53.5 $\pm$ 0.36 & 53.2 $\pm$ 0.28 & 52.7 $\pm$ 0.31 
\\
\midrule
\multirow{4}{*}{\centering GRU} 
 & A &  53.8 $\pm$ 0.43 & 52.9 $\pm$ 0.35 & 52.2 $\pm$ 0.43 & 51.5 $\pm$ 0.46 & 50.9 $\pm$ 0.42 
 \\
 & V &  51.6 $\pm$ 0.29 & 51.2 $\pm$ 0.31 & 51.1 $\pm$ 0.31 & 50.8 $\pm$ 0.28 & 50.6 $\pm$ 0.28 
 \\
 & A+V & 58.1 $\pm$ 0.24 & 57.2 $\pm$ 0.15 & 56.3 $\pm$ 0.23 & 55.4 $\pm$ 0.17 & 54.4 $\pm$ 0.15 
 \\
 & A+V\textsuperscript{P} & 53.8 $\pm$ 0.37 & 53.0 $\pm$ 0.40 & 52.4 $\pm$ 0.44 & 51.8 $\pm$ 0.34 & 51.4 $\pm$ 0.40 
 \\
\midrule
\multirow{4}{*}{\centering Mamba} 
 & A &  52.9 $\pm$ 0.50 & 52.0 $\pm$ 0.47 & 51.1 $\pm$ 0.43 & 50.2 $\pm$ 0.50 & 49.6 $\pm$ 0.47 
 \\
 & V &  50.7 $\pm$ 0.23 & 50.5 $\pm$ 0.26 & 50.4 $\pm$ 0.34 & 50.0 $\pm$ 0.31 & 49.7 $\pm$ 0.30 
 \\
 & A+V & 53.9 $\pm$ 0.36 & 53.5 $\pm$ 0.43 & 53.1 $\pm$ 0.46 & 52.3 $\pm$ 0.52 & 51.8 $\pm$ 0.43 
 \\
 & A+V\textsuperscript{P} & 52.7 $\pm$ 0.41 & 51.8 $\pm$ 0.51 & 51.4 $\pm$ 0.46 & 50.7 $\pm$ 0.44 & 50.1 $\pm$ 0.50 
 \\
\bottomrule
\end{tabular}
\end{adjustbox}
\label{tab:timesteps_different_seeds_EasyCom_b}
\caption*{(b) 5 Different Seeds Results on EasyCom - Timesteps from 1.20s to 2.00s} 
\end{subtable}

\caption{Per-frame performance over 5 different random seeds on EasyCom.}

\label{tab:main_result_error_bar_easycom}
\end{table*}

\begin{table*}[t]
\centering
\footnotesize

\begin{subtable}{\textwidth}
\centering
\begin{adjustbox}{width=\textwidth, center}
\begin{tabular}{llccccc}
\toprule
\multirow{2}{*}{Model} & \multirow{2}{*}{Modality} & \multicolumn{5}{c}{mAP (\%)} \\
\cmidrule(lr){3-7}
 & & 0.20s & 0.40s & 0.60s & 0.80s & 1.00s \\
\midrule
\multirow{4}{*}{\centering Transformer} 
 & A & 78.8 $\pm$ 0.06 & 74.9 $\pm$ 0.05 & 71.8 $\pm$ 0.04 & 69.7 $\pm$ 0.05 & 68.1 $\pm$ 0.02 
 \\
 & V & 58.7 $\pm$ 0.25 & 58.5 $\pm$ 0.25 & 58.4 $\pm$ 0.25 & 58.2 $\pm$ 0.24 & 58.1 $\pm$ 0.24 
 \\
 & A+V & 78.1 $\pm$ 0.20 & 74.3 $\pm$ 0.23 & 71.5 $\pm$ 0.24 & 69.4 $\pm$ 0.26 & 68.0 $\pm$ 0.27 
 \\
 & A+V\textsuperscript{P} & 78.4 $\pm$ 0.21 & 74.5 $\pm$ 0.17 & 71.5 $\pm$ 0.18 & 69.4 $\pm$ 0.20 & 67.9 $\pm$ 0.19 
 \\
\midrule
\multirow{4}{*}{\centering GRU} 
 & A & 78.6 $\pm$ 0.27 & 74.8 $\pm$ 0.30 & 71.8 $\pm$ 0.27 & 69.6 $\pm$ 0.27 & 68.1 $\pm$ 0.28 
 \\
 & V & 58.6 $\pm$ 0.59 & 58.3 $\pm$ 0.58 & 58.1 $\pm$ 0.58 & 57.9 $\pm$ 0.59 & 57.8 $\pm$ 0.61 
 \\
 & A+V & 76.4 $\pm$ 0.41 & 73.0 $\pm$ 0.42 & 70.4 $\pm$ 0.40 & 68.5 $\pm$ 0.39 & 67.1 $\pm$ 0.40 
 \\
 & A+V\textsuperscript{P} & 76.9 $\pm$ 0.18 & 73.4 $\pm$ 0.15 & 70.6 $\pm$ 0.15 & 68.6 $\pm$ 0.17 & 67.3 $\pm$ 0.19 
 \\
\midrule
\multirow{4}{*}{\centering Mamba} 
 & A & 77.4 $\pm$ 0.53 & 73.6 $\pm$ 0.44 & 70.5 $\pm$ 0.35 & 68.5 $\pm$ 0.35 & 66.9 $\pm$ 0.35 
 \\
 & V & 58.2 $\pm$ 0.29 & 58.1 $\pm$ 0.28 & 57.9 $\pm$ 0.29 & 57.8 $\pm$ 0.30 & 57.6 $\pm$ 0.29 
 \\
 & A+V & 76.0 $\pm$ 0.23 & 72.5 $\pm$ 0.22 & 69.8 $\pm$ 0.20 & 67.9 $\pm$ 0.18 & 66.6 $\pm$ 0.16 
 \\
 & A+V\textsuperscript{P} & 74.1 $\pm$ 0.38 & 70.8 $\pm$ 0.34 & 68.1 $\pm$ 0.34 & 66.2 $\pm$ 0.30 & 64.8 $\pm$ 0.33 
 \\
\bottomrule
\end{tabular}
\end{adjustbox}
\label{tab:timesteps_different_seeds_Ego4D_a}
\caption*{(a) 5 Different Seeds Results on Ego4D - Time Steps from 0.20s to 1.00s}
\end{subtable}

\begin{subtable}{\textwidth}
\centering
\begin{adjustbox}{width=\textwidth, center}
\begin{tabular}{llccccc}
\toprule
\multirow{2}{*}{Model} & \multirow{2}{*}{Modality} & \multicolumn{5}{c}{mAP (\%)} \\
\cmidrule(lr){3-7}
 & & 1.20s & 1.40s & 1.60s & 1.80s & 2.00s \\
\midrule
\multirow{4}{*}{\centering Transformer} 
 & A & 67.0 $\pm$ 0.03 & 66.3 $\pm$ 0.03 & 65.7 $\pm$ 0.04 & 65.1 $\pm$ 0.04 & 64.7 $\pm$ 0.04 
 \\
 & V & 58.0 $\pm$ 0.26 & 57.9 $\pm$ 0.25 & 57.8 $\pm$ 0.26 & 57.7 $\pm$ 0.24 & 57.7 $\pm$ 0.25 
 \\
 & A+V & 67.0 $\pm$ 0.26 & 66.3 $\pm$ 0.25 & 65.7 $\pm$ 0.25 & 65.3 $\pm$ 0.26 & 64.9 $\pm$ 0.27 
 \\
 & A+V\textsuperscript{P} & 66.7 $\pm$ 0.20 & 65.9 $\pm$ 0.20 & 65.4 $\pm$ 0.21 & 65.0 $\pm$ 0.21 & 64.5 $\pm$ 0.18 
\\
\midrule
\multirow{4}{*}{\centering GRU} 
 & A & 66.9 $\pm$ 0.24 & 66.2 $\pm$ 0.25 & 65.6 $\pm$ 0.25 & 65.2 $\pm$ 0.23 & 64.8 $\pm$ 0.26 
 \\
 & V & 57.8 $\pm$ 0.60 & 57.7 $\pm$ 0.61 & 57.6 $\pm$ 0.62 & 57.5 $\pm$ 0.64 & 57.5 $\pm$ 0.65 
 \\
 & A+V & 66.3 $\pm$ 0.39 & 65.6 $\pm$ 0.45 & 65.2 $\pm$ 0.50 & 64.7 $\pm$ 0.47 & 64.4 $\pm$ 0.44 
 \\
 & A+V\textsuperscript{P} & 66.3 $\pm$ 0.21 & 65.6 $\pm$ 0.21 & 65.1 $\pm$ 0.24 & 64.7 $\pm$ 0.22 & 64.4 $\pm$ 0.28 
 \\
\midrule
\multirow{4}{*}{\centering Mamba} 
 & A & 65.8 $\pm$ 0.35 & 65.0 $\pm$ 0.35 & 64.3 $\pm$ 0.33 & 63.9 $\pm$ 0.36 & 63.5 $\pm$ 0.37 
 \\
 & V & 57.5 $\pm$ 0.27 & 57.5 $\pm$ 0.28 & 57.4 $\pm$ 0.29 & 57.4 $\pm$ 0.29 & 57.3 $\pm$ 0.28 
 \\
 & A+V & 65.6 $\pm$ 0.17 & 64.8 $\pm$ 0.17 & 64.2 $\pm$ 0.19 & 63.8 $\pm$ 0.22 & 63.5 $\pm$ 0.22 
 \\
 & A+V\textsuperscript{P} & 63.9 $\pm$ 0.27 & 63.2 $\pm$ 0.19 & 62.7 $\pm$ 0.18 & 62.3 $\pm$ 0.09 & 62.0 $\pm$ 0.11 
 \\
\bottomrule
\end{tabular}
\end{adjustbox}
\label{tab:timesteps_different_seeds_Ego4D_b}
\caption*{(b) 5 Different Seeds Results on Ego4D - Time Steps from 1.20s to 2.00s}
\end{subtable}

\caption{Per-frame performance over 5 different random seeds on Ego4D.}

\label{tab:main_result_error_bar_ego4d}
\end{table*}

\section{Use of AI Assistants}
We used Claude 3.5 Sonnet to revise the paper and code, and GitHub Copilot to write the code.

\end{document}